\documentclass[10pt,twocolumn,letterpaper]{article}

\usepackage{cvpr} %

\usepackage{float}
\usepackage{multirow}
\usepackage{float}      %
\usepackage{afterpage}
\usepackage{scalerel,xparse}
\usepackage{capt-of,etoolbox}
\definecolor{cvprblue}{rgb}{0.21,0.49,0.74}
\usepackage[pagebackref,breaklinks,colorlinks,allcolors=cvprblue]{hyperref}

\def\paperID{000} %
\def\confName{CVPR}
\def\confYear{2025}

\title{\method: Latent Intrinsics Meets Diffusion Models for Indoor Scene Relighting}

\vspace{-40pt}
\author{\hspace{7mm} Xiaoyan Xing$^{1}$~~~ Konrad Groh$^{2}$~~~ Sezer Karaoglu$^{1}$~~~ Theo Gevers$^{1}$~~~ Anand Bhattad$^{3}$~ \hspace{7mm}\\\vspace{-8pt}\\
~~~$^{1}$University of Amsterdam~~~ $^{2}$BCAI-Bosch~~~ $^{3}$Toyota Technological Institute at Chicago  \\\vspace{-10pt}\\
\normalfont{\url{https://luminet-relight.github.io}}
}
\newcommand{\method}{{LumiNet}\xspace}
\begin{document}
\makeatletter
\g@addto@macro\@maketitle{
	\begin{figure}[H]
	\scriptsize
		\setlength{\linewidth}{\textwidth}
		\setlength{\hsize}{\textwidth}
		\vspace{-8.5mm}
  \centering
  \footnotesize
  \setlength\tabcolsep{0.2pt}
  \renewcommand{\arraystretch}{0.1}
\includegraphics[width=\textwidth]{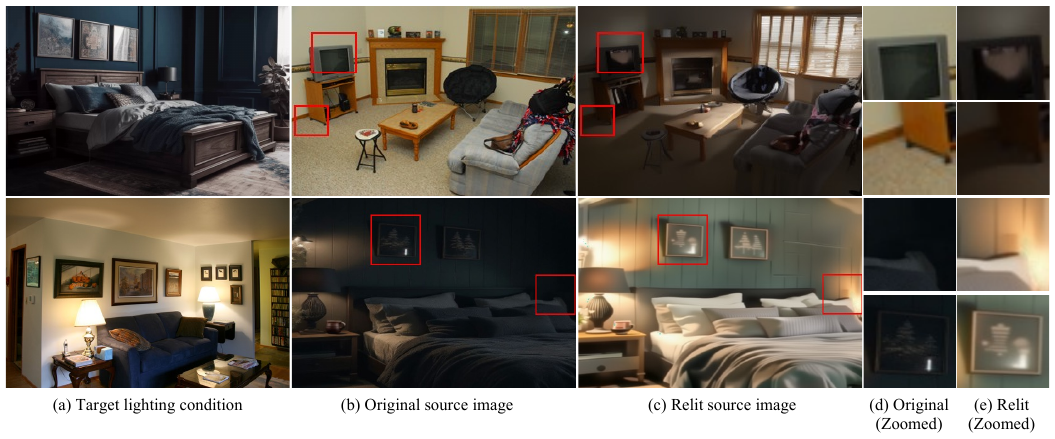}
\vspace{-15pt}
    \caption{\method transfers complex lighting conditions from a target image (a) to a source image (b), synthesizing a relit version of the source image (c) while preserving its geometry and albedo. In the top row, observe how \method transforms the scene from nighttime to daytime by transferring strong directional light from the target image’s window to the source image. Key details in the relit image include pronounced gloss on the table, shadows cast onto the carpet (center left), cast shadows from the TV stand (left corner), and, most importantly, reflections of the table on the TV screen (d, e). These changes demonstrate plausible control over both direct and indirect lighting effects, such as reflections, specular highlights and shadow placement. In the bottom row, \method ``knows'' about luminaires. In the relit image, two bedside lamps illuminate the scene, transforming it from a dimly lit room into a well-lit environment. This suggests that \method recognizes the spatial arrangement of objects and infers where light sources should be switched on. Note how \method introduces specular highlights on the left painting (see crop) and gloss in the far-right corner of the bedroom, where a previously invisible bedside lamp is now turned on. These results show \method's ability to handle complex lighting phenomena—including direct illumination, specular highlights, cast shadows, inter-reflections and other indirect effects—while maintaining scene geometry, and albedo.
    }
        \label{fig:teaser}
	\end{figure}
}
\makeatother    
\maketitle
\begin{abstract}
We introduce \method, a novel architecture that leverages generative models and latent intrinsic representations for transferring lighting from one image to another. Given a source image and a target lighting image, \method generates a relit version of the source scene that captures the target's lighting.  Our approach makes two key contributions: a data curation strategy from the StyleGAN-based relighting model for our training, and a modified diffusion-based ControlNet that processes both latent intrinsic properties from the source image and latent extrinsic properties from the target image. We further improve lighting transfer through a learned adaptor that injects the target's latent extrinsic properties via cross-attention and light-weight fine-tuning. 

Unlike traditional ControlNet, which generates images with conditional maps from a single scene, \method processes latent representations from two different images - preserving geometry and albedo from the source while transferring lighting characteristics from the target. Experiments demonstrate that our method successfully transfers complex lighting phenomena including specular highlights and indirect illumination across scenes with varying spatial layouts and materials, outperforming existing approaches on challenging indoor scenes using only images as input.

\end{abstract}
    
\section{Introduction}
\label{sec:intro}
Transferring lighting conditions between indoor scenes has applications in cinematography, architectural visualization, and mixed reality. While recent advances in neural rendering have shown promising results for single image relighting, transferring lighting between different images remains challenging due to the complex interplay of scene geometry, materials, and illumination.

The key challenge stems from the difficulty in decomposing and transferring lighting effects between scenes with different spatial layouts and surface properties. Moreover, light in scenes cannot just appear but must come from luminaires, meaning
that transferring a lighting pattern from scene to scene requires a detailed
understanding of light sources in the scene. Furthermore, indoor scenes have complex light transport phenomena including interreflections, shadows, and spatially-varying material interactions that are highly scene-specific~\cite{zhang2021neural}. Traditional inverse rendering approaches attempting to recover scene components explicitly often struggle with model limitations and error propagation~\cite{li2022physically}. Other approaches either require extensive multi-view capture setups, are limited to specific object categories~\cite{Neural-Garffer, zeng2024dilightnet} or portraits~\cite{ponglertnapakorn2023difareli, kim2024switchlight}, or cannot transfer complex lighting effects between different scenes \cite{zeng2024rgb,Latent-Intrinsic}.

Recent studies have shown promising directions. \citet{StyLitGAN} showed that StyleGAN's latent space \cite{karras2019_stylegan} contains disentangled lighting representations and uses them to manipulate the lighting of generated images, but their approach does not transfer well to real images~\cite{bhattad2023make}. Zhang et al. demonstrated that latent intrinsic decomposition can capture emergent properties of albedo and illumination, and can be used for relighting~\cite{Latent-Intrinsic}. While these representations are robust, our experiments demonstrate they do not generalize to complex, arbitrary scenes. Meanwhile, diffusion models~\cite{ho2020denoising,rombach2022high_latentdiffusion_ldm} with ControlNet~\cite{zhang2023adding_controlnet} have shown remarkable conditional image-generation capabilities. 
DiffusionLight~\cite{Phongthawee2023DiffusionLight} recovers environment maps by inpainting chrome balls, while IC-Light~\cite{IC-Light} relights portrait images. However, these methods cannot relight complex indoor scenes.

We present \method, a novel approach that synthesizes the strengths of these different generators while addressing their individual limitations. Our key insight is that by carefully modifying the ControlNet architecture to operate on latent representations of scene intrinsics and extrinsics~\cite{Latent-Intrinsic}, we can achieve robust lighting transfer between arbitrary indoor scenes. First, we develop a training pipeline that integrates a variational StyleGAN architecture with real indoor scene data to alleviate mode collapse issues common in indoor scene generation. This approach also addresses the lack of training data for real indoor scenes lit under different lighting conditions. Second, we train a \emph{Latent ControlNet} that learns to decompose and transfer lighting features by operating in learned latent spaces and using lighting feature-aware fine-tuning, without requiring explicit 3D reconstruction or material modeling. Third, we introduce a lighting-aware adaptor network that maps a low-dimensional latent lighting extrinsic vector to a high-dimensional code. This code is injected into a pretrained diffusion model by fine-tuning its cross-attention layers, helping the model to preserve target lighting characteristics effectively.

Our method successfully relights challenging cases where the target (\cref{fig:teaser}a) and the source images (\cref{fig:teaser}b) differ significantly in spatial arrangements and material properties, exploiting learned priors from powerful image generators. Results (\cref{fig:teaser}c) demonstrate that our relighting method can create complex lighting phenomena in physically plausible ways, including specular highlights, soft shadows, and indirect illumination effects like inter-reflections (as shown in \cref{fig:teaser}d; see the TV in the top row). Extensive experiments demonstrate that \method outperforms previous methods by requiring only images as input, while adeptly preserving complex lighting effects, particularly on in-the-wild internet images.

In summary, our main contributions are:
\begin{itemize}
    \item Novel Framework: \method combines latent intrinsic control with diffusion models for high-quality indoor scene relighting without 3D or multi-view inputs.
    \item Training Data: A variational StyleGAN approach maps real images to latent space of StyleGAN, enabling diverse data generation for our training.
    \item Generalizable Relighting: Despite training only on same-scene pairs, \method successfully transfers lighting between scenes in the wild with different layouts.
    \item Plausible Lighting Effects: \method can relight diverse indoor scenes with complex lighting effects, including specular highlights, cast shadows, and inter-reflections. Extensive evaluations (quantitative and qualitative) and user studies validate \method's effectiveness.
\end{itemize}

\section{Related Work}
\begin{figure*}[t]
    \centering
\includegraphics[width=\linewidth]{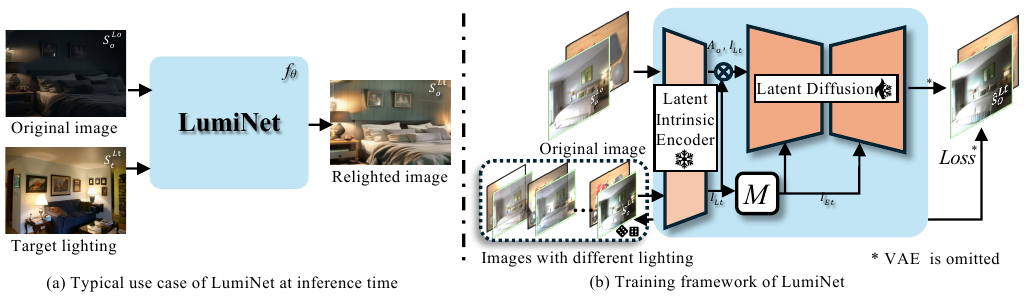}
\vspace{-20pt}
 \caption{\textbf{\method's Architecture and Training Pipeline.} \textbf{Left}: \method's inference pipeline transfers lighting from a target image to a source image while preserving structure and materials (a). \textbf{Right}: Training utilizes latent intrinsic representations from a pretrained model~\cite{Latent-Intrinsic}, which decomposes images into lighting-invariant features and low-dimensional extrinsic lighting vectors. We train a conditional latent diffusion model with a lightweight MLP adaptor $M$ to align lighting extrinsics with text embedding dimensions, using empty prompts for conditioning. Paired scenes with varying lighting guide training via latent diffusion loss (VAE encoder/decoder omitted). Despite training on same-scene relight pairs, \method generalizes well to scenes with different layouts and materials.}
    \label{fig:pipeline}
    \vspace{-10pt}
\end{figure*}
\label{sec:related work}
Transferring lighting conditions across scenes requires understanding each scene's intrinsic properties and lighting. We review related work by categorizing methods based on their reliance on intrinsic images or direct image inputs.

\vspace{-10pt}
\paragraph{Intrinsic Image Relighting.}
Intrinsic image methods explicitly rely on scene intrinsics, particularly in challenging indoor scenarios involving complex lighting interactions. Traditional approaches often use detailed 3D reconstruction and inverse graphics, which are computationally intensive~\cite{indoor_3d_1, indoor_3d_2, indoor_3d_3, li2022physically, indoor_3d_5}. Other methods leverage intrinsic images primarily for object-level compositing or relighting, maintaining consistency with background lighting but not addressing full-scene relighting~\cite{bhattad2022cut, careaga2023harmonization, zhang2024zerocomp, fortier2024spotlight, liang2024photorealistic}. Methods that are multi-view based further provide richer information about the scene as demonstrated in outdoor time-lapse relighting~\cite{duchene2015multi, liu2020learning}, geometry-aware networks~\cite{philip2019multi}, neural radiance fields for objects~\cite{zhang2021nerfactor, Toschi_2023_CVPR}, outdoor scenes~\cite{gao2023relightable, rudnev2022nerfosr, gardner2024sky, lin2023urbanir}, and human portraits~\cite{cai2024real, NeRFFaceLighting}.

Single image methods have advanced by 
conditioning generative diffusion models~\cite{rombach2022high_latentdiffusion_ldm} on shading, normals, and intrinsic representations~\cite{zeng2024rgb, kocsis2024lightit, luo2024intrinsicdiffusion, kocsis2024iid, choi2025scribblelight, zhang2024zerocomp, fortier2024spotlight}. Despite their interpretability, these explicit methods face limitations in intrinsic prediction accuracy and complex real-world lighting scenarios. Our approach addresses these challenges by utilizing latent intrinsic representations~\cite{Latent-Intrinsic} directly extracted from complex single image as inputs.

\vspace{-10pt}
\paragraph{Image-based Relighting.}
Image-based relighting has notably progressed in specialized domains, including portrait relighting~\cite{sun2019single, zhou2019deep, nestmeyerlearning, sengupta2021light, ponglertnapakorn2023difareli, kim2024switchlight, scribble_relit} and outdoor scenes with simpler illumination conditions~\cite{yu2020self, liu2020learning}. However, generalizing to diverse real-world scenes remains challenging, despite initial successes with self-attention~\cite{hu2020sa} and depth guidance~\cite{yang2021s3net}. StyLitGAN~\cite{bhattad2024stylegan} explored latent lighting representations but was limited to synthetic images, and its extension using GAN inversion~\cite{bhattad2023make} struggles to generalize to real-world images.

Recent diffusion-based approaches, including DilightNet~\cite{zeng2024dilightnet}, IllumiNeRF~\cite{IllumiNeRF}, Neural Gaffer~\cite{Neural-Garffer}, and FlashTex~\cite{FlashTex}, primarily focus on object-level relighting via light-aware ControlNets. \citet{GS-MIIW} achieve multi-view relighting using direct multi-lighting data~\cite{MIIW} and Gaussian splatting~\cite{kerbl3Dgaussians}. Retinex-diffusion~\cite{xing2024retinex} proposes a training-free diffusion scheme based on retinex theory~\cite{land1977retinex}, yet is constrained by predefined lighting directions. IC-Light~\cite{IC-Light} relights foreground convincingly but struggles with full-scene relighting. In contrast, our method relights full-scene outperforming recent approaches including IC-Light~\cite{IC-Light} and RGB$\leftrightarrow$X~\cite{zeng2024rgb}.

\paragraph{Intrinsic Image Decomposition}
traces back to Barrow and Tenenbaum~\cite{barrowtenenbaum}. Early methods like SIRFS~\citep{barron2015shape}, use shading information to recover shape, illumination, and reflectance. Comprehensive reviews are available in~\cite{forsyth2021intrinsic, garces2022survey}. Recent works utilize synthetic training data to improve intrinsic estimations via ordinal shading~\cite{careaga2023intrinsic, careaga2024colorful, dilleIntrinsicHDR}, surface normals~\cite{baslamisli2021shadingnet}, color priors~\cite{das2022pie}, and point cloud representations~\cite{xing2023intrinsic}. Notably, Bhattad et al.~\cite{bhattad2024stylegan} and Du et al.~\cite{du2023generative} demonstrated that intrinsic images naturally emerge within generative models. Conditional generative models further improve intrinsic estimation using diffusion priors~\cite{kocsis2024iid, zeng2024rgb, luo2024intrinsicdiffusion, chen2024intrinsicanything}.

An alternative, recent paradigm treats intrinsic representations as latent variables~\cite{Latent-Intrinsic}, revealing latent albedo-like representations emerge with simplified lighting control. We extend this latent intrinsic approach by projecting these representations into generative diffusion model's latent spaces, resulting in flexible and improved lighting manipulation.

\section{Overview}  
As illustrated in \cref{fig:pipeline}, given a real-world scene \( S_o \) with lighting \( L_o \), we learn a model \( f_\theta \) to replace \( L_o \) with the lighting \( L_t \) from a target scene \( S_t \):  

\begin{equation}  
\label{equ:relit_model}  
    f_\theta(S_o^{L_o}, S_t^{L_t}) \rightarrow S_o^{L_t}.  
\end{equation}  

\begin{figure*}[ht!]
    \centering
    \includegraphics[width=\linewidth]{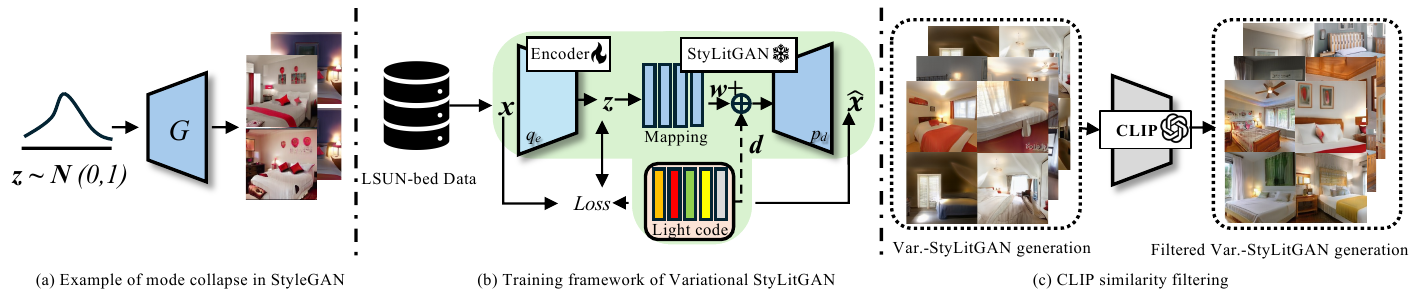}
    \vspace{-18pt}

    \caption{\textbf{Training Framework of Variational StyLitGAN.}  
    (a) Traditional StyleGAN suffers from mode collapse, generating similar outputs every 10–20 iterations despite different latent codes.  
(b) Our variational approach maps real images to StyleGAN's latent space via an encoder (\( q_e \)), using a frozen pretrained generator (\( p_g \)) from StyLitGAN~\cite{StyLitGAN}. The colored bars represent StyLitGAN’s disentangled lighting codes, enabling diverse scene generation under varying lighting. While the mapping is approximate, it provides sufficient diversity for training LUMINET by leveraging real-image variations.  
(c) We apply CLIP similarity filtering to ensure high-quality samples.  }
    \label{fig:VAE-GAN}
    \vspace{-15pt}
\end{figure*}

Traditional methods decompose and re-render lighting using inverse rendering and ray tracing. In contrast, we frame relighting as a conditional image generation task, conditioning on intrinsic properties of the scene image we aim to relight and a low-dimensional target lighting representation extracted from another image whose lighting we wish to match. Previous generative approaches rely on image-space representations such as environment maps, which struggle with scene-level relighting since they fail to capture internal light sources and remain closely tied to scene geometry.

We introduce a latent feature-based approach jointly representing scene properties and target lighting, enabling end-to-end relighting (\cref{fig:pipeline}). However, effective relighting requires: (1) Intrinsic consistency, preserving scene structure and materials; (2) Plausible lighting transfer, ensuring the lighting appears natural. To achieve this, we utilize a carefully constructed dataset (Sec.\ref{Sec: Data}), extending StylitGAN\cite{StyLitGAN}, and a systematic approach to lighting transfer (Sec.\ref{Sec:method}) that leverages latent intrinsic representations\cite{Latent-Intrinsic}. %

\section{Data Preparation}
\label{Sec: Data}
Acquiring paired images of real-world scenes under different lighting conditions is extremely challenging, requiring carefully controlled environments and extensive setups. To address this limitation, we develop a two-stage data preparation strategy: (1) a variational synthetic scene generation approach that captures essential lighting patterns, and (2) a curated collection of in-the-wild images that ensures diverse and balanced training data. This combination enables our model to learn robust lighting transfer while maintaining photorealistic quality.
\subsection{Variational Relit Scene Generation}

StyLitGAN~\cite{StyLitGAN} generates plausible relit images by interpolating StyleGAN's latent space. It maps random Gaussian noise $\mathbf{z}$ to a latent style code $\mathbf{w}$, then adds a predefined lighting direction $\mathbf{d}$ to generate relit images.

However, StyleGAN~\cite{karras2019_stylegan} can suffer from mode collapse when exploring its high-dimensional latent space, producing nearly identical outputs from different samples of the Gaussian distribution (\cref{fig:VAE-GAN}(a)). One workaround is to map real images into the latent space via GAN inversion. However, the best-performing inversion method~\cite{bhattad2023make} uses an optimization-based approach, which is computationally slow and unsuitable for large-scale data curation.

To address this, we propose variational-StyLitGAN (\cref{fig:VAE-GAN}(b)), which maps real-world images into StyleGAN's latent space using a ConvNeXt-based~\cite{ConvNext} variational encoder $q_{e}(\mathbf{z}|\mathbf{x})$. The encoder maps an input image $\mathbf{x}$ to a variational latent code $\mathbf{z}$, which is then transformed into a style code $\mathbf{w^{+}}$ by a pretrained mapper. A frozen StyLitGAN generator $p_{d}(\mathbf{x}|\mathbf{w^{+}})$ reconstructs the scene image $\hat{\mathbf{x}}$.

We optimize the network using:
\begin{equation}
\begin{aligned}
\mathcal{L} = & \underbrace{\text{MSE}(\mathbf{x}, \hat{\mathbf{x}}) + \text{LPIPS}(\mathbf{x}, \hat{\mathbf{x}})}_{\mathcal{L}_{\text{rec}}} \\
              & + \underbrace{D_{\text{KL}}(q_{\phi}(\mathbf{z}|\mathbf{x}) \parallel \mathcal{N}(0, I))}_{\mathcal{L}_{\text{KL}}}
\end{aligned}
\end{equation}
where $\mathcal{L}_{\text{rec}}$ combines MSE and perceptual loss (LPIPS)~\cite{LPIPS} for accurate reconstruction, and $\mathcal{L}_{\text{KL}}$ regularizes the latent distribution.

For dataset generation, we encode LSUN-bedroom images to obtain $\mathbf{z}$, map to $\mathbf{w^{+}}$, and add lighting direction $\mathbf{d}$ to generate seven lighting variations per scene. We further curate $\approx$1K high-quality unique images using CLIP~\cite{radford2021learning} similarity to the keywords photo-realistic'', good lighting'', and ``illumination'' (\cref{fig:VAE-GAN}(c)).

While StyLitGAN provides good lighting control for generated images, the gap between generated and real images makes it challenging to train solely on synthetic data. Therefore, we use this pipeline primarily for data generation, leveraging its diverse lighting variations to train \method for cross-scene light transfer.

\subsection{In-the-Wild Training Data}
To complement our generated samples, we leverage several real-world datasets:
Multi-Illumination Images in the Wild (MIIW)\cite{MIIW} provides controlled lighting variations across over 1,000 indoor scenes, each captured under 25 distinct conditions, offering high-quality specular effects and direct lighting. BigTime\cite{Bigtime} contributes diverse lighting effects including hard shadows through time-lapse captures of 460 scenes under 20--50 lighting conditions. We additionally sample 1,000 images per training from LSUN Bedroom~\cite{LSUN} to enhance training distribution diversity.

Unlike prior works focused on object-level or portrait relighting~\cite{Neural-Garffer, ponglertnapakorn2023difareli, IC-Light}, our approach targets scene-level relighting, thus avoiding object-centric datasets.

Summing it up, we train \method on $\sim$2,500 unique scenes with their relit pairs and 1,000 scenes from LSUN for which we do not have relighting pairs.

\section{\method}
\label{Sec:method}
Our goal is to learn a generative model that can transfer lighting between indoor scenes while preserving scene structure. The key challenge lies in conceptualizing lighting and its complex interactions within scenes. Our solution leverages latent intrinsic representations during training, grounded in image formation theory which separates images into illumination-invariant (intrinsic) and illumination-dependent (extrinsic) components.

\subsection{Latent Intrinsic Extraction}
Traditional intrinsic decomposition in pixel space (e.g., albedo, roughness, surface normals) faces two key challenges: (1) perfect decomposition from monocular images is nearly impossible, and (2) obtaining all necessary components is computationally expensive. Instead, we process intrinsic information entirely in latent space.

Building on pretrained model from \citet{Latent-Intrinsic}, given an image pair $(S_o^{L_o}, S_o^{L_t})$ of scene $S_o$ under different lighting conditions $L_o$ and $L_t$, we use a pre-trained latent-intrinsic encoder $f_{\lambda}$ to extract latent intrinsic features $\mathcal{A}_{o} \in \mathbb{R}^{H \times W \times 128}$ and lighting codes $\{\mathcal{I}_{L_o}, \mathcal{I}_{L_t}\} \in \mathbb{R}^{16}$.

\subsection{Latent Intrinsic Control}
Our illumination control scheme consists of two key components. First, unlike traditional ControlNet \cite{zhang2023adding_controlnet} that operates on images, we implement control directly in latent space through our Latent Intrinsic ControlNet. We expand the target latent illumination $\mathcal{I}_{L_t}$ to match spatial dimensions of $\mathcal{A}_{o}$, then concatenate them to form $\{\mathcal{A}_{o},\mathcal{I}_{L_t'}\} \in \mathbb{R}^{H \times W \times 144}$. This concatenated feature is processed through convolution layers to obtain $\mathcal{L}\in \mathbb{R}^{H/2\times W/2\times 512}$.

Second, we enhance lighting control through cross-attention in the diffusion model. A learned MLP ($3072\rightarrow4096\rightarrow4096\rightarrow4096\rightarrow3072$) transforms the lighting code (with necessary rescaling) into $\mathcal{I}_{E_t}\in \mathbb{R}^{3 \times 1024}$, matching the text embedding dimensions. We exclude text prompts to focus purely on image-based lighting transfer.

\subsection{Training Objective}
During training, we focus on same-scene lighting transfer through a latent diffusion process. The process begins by encoding target lighting scene $S^{L_t}$ to latent $\epsilon(S^{L_t})$, then progressively adds noise to obtain $\epsilon(S^{L_t})_t$. The model predicts noise using multiple conditions: time step $t$, latent features $\{\mathcal{A}_{o}, \mathcal{I}_{L_t'}\}$, lighting embedding $\mathcal{I}_{E_t}$ and original scene $S^{L_o}$. The objective function is:

\begin{equation}
    \mathcal{L}_{\text{Lumi}} = \|\epsilon - \theta(\epsilon(S^{L_t})_t, t, \{\mathcal{A}_{o}, \mathcal{I}_{L_t'}\},\mathcal{I}_{E_t},\epsilon(S^{L_o}))\|^2_2
\end{equation}

We train only the latent control network and cross-attention layers while keeping other diffusion model and latent intrinsic model parameters frozen.

\subsection{Bypass-decoder}
Since lighting control occurs in the latent space, the default autoencoder in Stable Diffusion may be limited in its ability to accurately reconstruct fine details of the input image, which could lead to issues in preserving geometry details. Inspired by earlier work in low-light image enhancement \cite{quadprior}, which suggests using a bypass decoder to improve the visual quality of generated images, we adopt a similar approach. Specifically, we replace the default decoder during inference with a lightweight, fine-tuned bypass decoder.

This bypass decoder is fine-tuned on the MIIW and BigTime datasets. Following a setup similar to \cite{quadprior}, we apply random color jittering and additive noise to the input images during training to help the decoder better preserve the original image's identity.

\subsection{Post-processing}
Given the inherent randomness in the denoising process of diffusion models, we introduce two post-processing techniques for real-world scene relighting. Notably, no post-processing is applied to the MIIW relighting results.

\noindent\textbf{Nearest Neighbor based Selection.}
Despite \method's generalizable ability in light transfer, the generative model is still affected by initial seeds~\cite{xu2024good}, which can produce sub-optimal relighting results, particularly when precise control over local lighting effects is required, such as turning lamps on and off. We propose a nearest neighbor searching scheme based on the latent lighting code of images generated with random seeds and the target lighting image. 

\noindent\textbf{Flow-Based Clean Up.}
While our method performs well for conditioned relighting effects, a U-Net-based diffusion model may still produce sub-optimal artifacts in complex indoor scenes. We employ rectified-flow inversion \cite{rout2024rfinversion} with $\eta=0.99$ to remove artifacts and achieve higher resolution. Importantly, we do not introduce any prompts related to the lighting conditions of the image, to prevent any lighting-related changes by the rectified-flow model. 

\section{Experiment}
\label{Sec:Exp}
\begin{figure*}[t]
    \centering
\includegraphics[width=\linewidth]{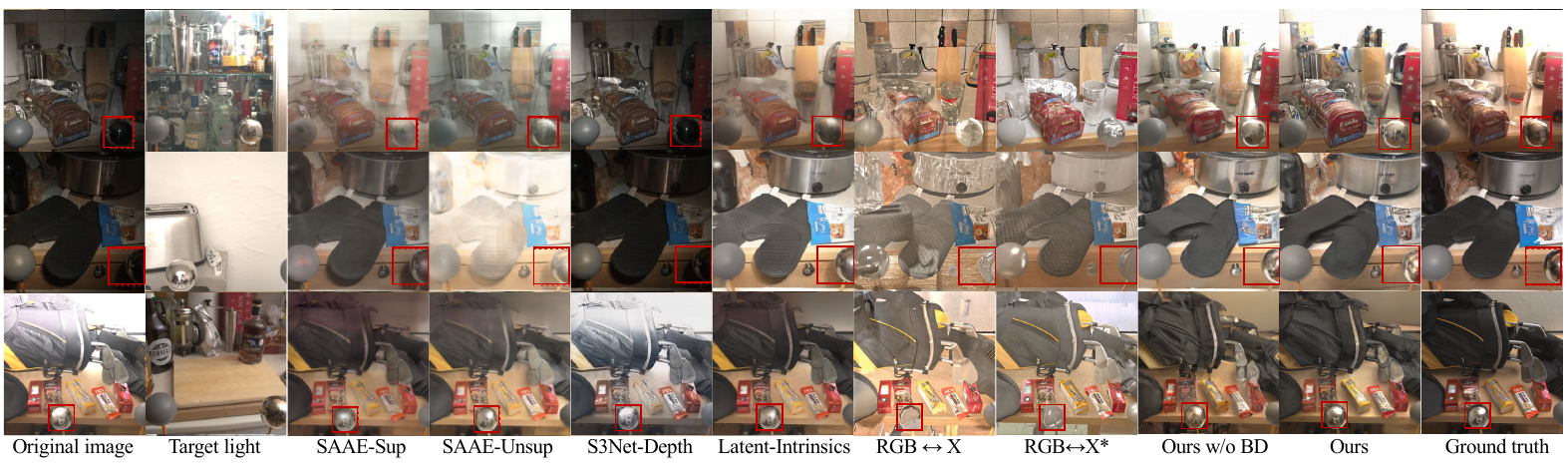}
\vspace{-20pt}
\caption{\textbf{Qualitative comparison on the MIIW dataset~\cite{MIIW}}. 
To facilitate a direct comparison with prior work, we evaluate on the same images from the MIIW benchmark as those reported in Latent-Intrinsics~\cite{Latent-Intrinsic}. Our method produces more realistic and consistent lighting effects. In contrast to Latent-Intrinsics, which fails at producing accurate cast shadows and highlights, and RGB$\leftrightarrow$X~\cite{zeng2024rgb}, which struggles with complex materials, our approach generates more plausible lighting. While the base version of our method shows minor geometric artifacts (Ours w/o BD), our final model (Ours), which incorporates a bypass decoder~\cite{quadprior}, best preserves the object's identity while generating superior lighting effects.}
    \vspace{-5pt}
    \label{fig:mit relight}
\end{figure*}

\begin{figure*}[t]
    \centering
\includegraphics[width=\linewidth]{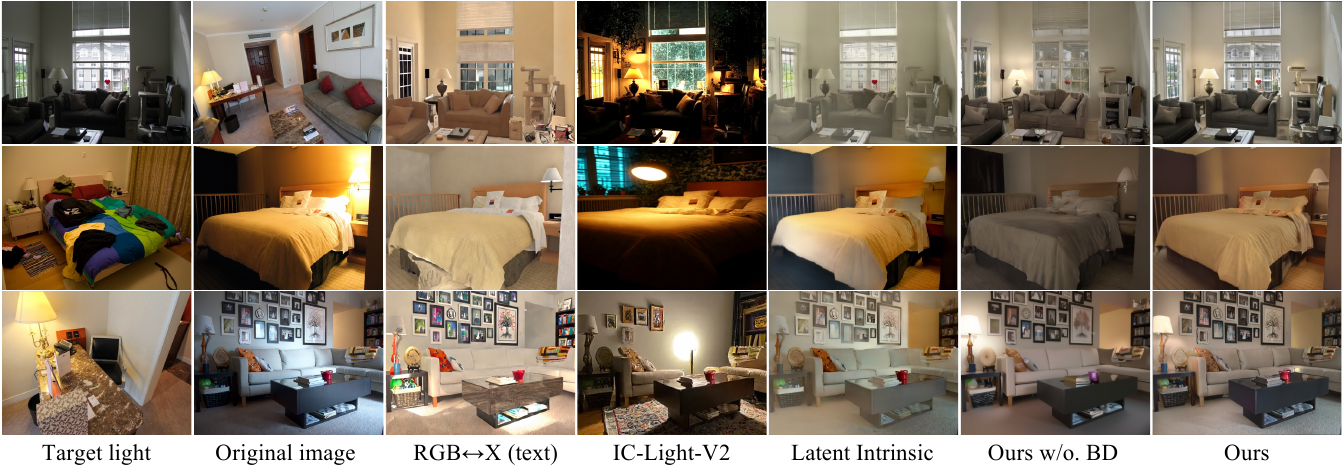}
\vspace{-20pt}
    \caption{\textbf{In-the-wild image relighting visual comparison.} We evaluate \method on diverse indoor scenes under various target lighting conditions, more in the supplemental. Both RGB$\leftrightarrow$X \cite{zeng2024rgb} and IC-Light-v2 \cite{IC-Light} require text prompts to achieve relighting, where we use descriptions derived from the target lighting image (including actions like turning lights on/off, lamp placement, and scene type) as text prompts. In contrast, Latent Intrinsic \cite{Latent-Intrinsic} and our method rely solely on image input. Since Latent Intrinsics is exclusively trained on MIIW images, it struggles to generalize to real-world, in-the-wild internet images. In contrast, \method effectively identifies the locations of luminaries and successfully transfers lighting, even when significant spatial layout changes occur.}
    \vspace{-5pt}
    \label{fig:all_result}
\end{figure*}

We evaluate the light transfer capability on both a controlled lighting dataset and real-world images both qualitatively (\cref{fig:mit relight} \& \cref{fig:all_result}) and quantitatively (\cref{tab:MIIW}). Additionally, we assess perceptual quality through a user study (\cref{tab:user_study}). %

\subsection{Quantitative Evaluation}
\label{Sec:QE}

We compare our method against recent advancements using deep networks (SA-AE \cite{hu2020sa}, S3Net, \cite{Latent-Intrinsic}) and diffusion models (RGB$\leftrightarrow$X~\cite{zeng2024rgb}) on the test set from the MIIW dataset, which was not included in our training set.

{We employ the official Latent-intrinsics \cite{Latent-Intrinsic} evaluation code to evaluate every method, specifically, to match the latent intrinsics' evaluation protocol, RMSE and SSIM are evaluated based on data range from [-1,1].}

As shown in ~\cref{tab:MIIW}, although our method does not directly surpass Latent-Intrinsic~\cite{Latent-Intrinsic} and SA-AE~\cite{hu2020sa}—both specifically trained for relighting tasks on the MIIW dataset—we outperform our closest diffusion-based counterpart, RGB$\leftrightarrow$X, which relies on G-buffer information for relighting. It is also worth noting that the current evaluation metrics are limited in their ability to accurately reflect relighting performance, as small pixel-level differences in highlight regions can lead to disproportionately large numerical errors \cite{giroux2024towards}.

In \cref{fig:mit relight}, we provide qualitative relighting results on the MIIW dataset. For this comparison, we evaluate our method on the same input and reference lighting image from the Latent-Intrinsics paper, and compare the results of RGB$\leftrightarrow$X and our method under identical conditions. Latent-Intrinsic struggles with challenging lighting scenarios, such as removing cast shadows, recovering overexposed regions, or generating realistic highlights on plastic surfaces. Meanwhile, RGB$\leftrightarrow$X exhibits limited capability in cross-scene lighting transfer. Although the variant RGB$\leftrightarrow$X*—which uses ground-truth irradiance—produces more meaningful relighting, it still fails to handle highly ambiguous materials like metal and glass. Our method effectively captures a wider range of lighting effects compared to other approaches; however, it can exhibit some fine geometric inconsistencies. With the inclusion of the bypass decoder (BD) from \cite{quadprior}, we observe that our approach significantly improves identity preservation while retaining these vivid and realistic lighting effects.

\begin{figure*}[t!]
    \centering
    \includegraphics[width=\linewidth]{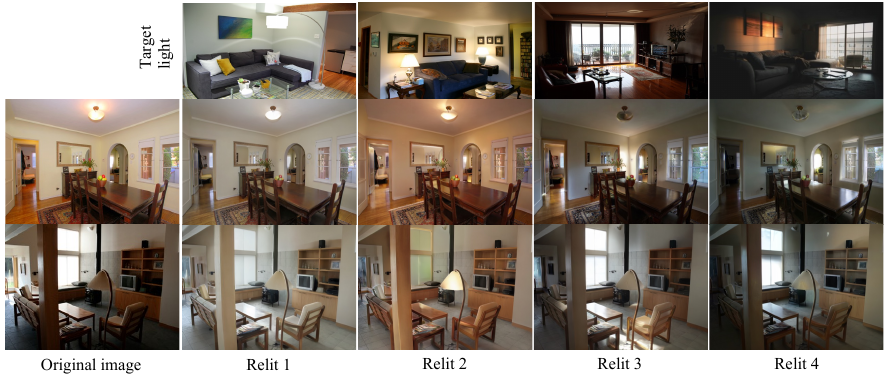}
    \vspace{-20pt}
    \caption{\textbf{Same Scene Under Various Lighting. }\method can relight the same scene under different lighting conditions while preserving the overall layout, demonstrating effective disentanglement of intrinsic properties and lighting. }
    \vspace{-5pt}
    \label{fig:m_relits}
\end{figure*}

\subsection{Geometry Consistency and User Study}

In open-world relighting scenarios, we evaluate our method based on surface normal consistency and conduct a user study to assess perceptual image relighting quality, as no ground truth is available. We compare \method (including w/o BD) with IC-Light~\cite{IC-Light}, RGB$\leftrightarrow$X~\cite{zeng2024rgb}, and Latent-Intrinsic \cite{Latent-Intrinsic}. The visual examples are in \cref{fig:all_result}. For RGB$\leftrightarrow$X~\cite{zeng2024rgb}, we use only the text-prompting relighting for the user study, as irradiance-based relighting is not effective in this setting (see Sec.~\ref{Sec:QE} for details). For IC-Light~\cite{IC-Light}, we use the latest FLUX version, specifically IC-Light-v2 (with foreground conditioning), as it offers the best performance. \cref{fig:m_relits} shows the same image relit using different reference images, demonstrating the usefulness of our relighting framework and robustness of our latent intrinsic and lighting disentanglement.

\begin{table}[t!]
    \centering
    \resizebox{\linewidth}{!}{\begin{tabular}[t]{lc|cc|cc}
    \toprule
\textbf{Methods} & \textbf{Labels} & \multicolumn{2}{c|}{\textbf{Raw Output}} & \multicolumn{2}{c}{\textbf{Color Correction}} \\
\midrule
\multicolumn{2}{c}{{Data Range [-1,1]}}& RMSE$\downarrow$ & SSIM$\uparrow$ & RMSE$\downarrow$ & SSIM$\uparrow$ \\
\midrule
Input Img & - & 0.384 & 0.438 & 0.312 & 0.492\\
\midrule
\multicolumn{6}{l}{\textit{Trained only on MIIW dataset}} \\
SA-AE~\cite{hu2020sa} & Light & \textbf{{0.288}} & \textbf{{0.484}} & 0.232 & 0.559\\
SA-AE~\cite{hu2020sa} & - & 0.443 & 0.300 & 0.317 & 0.431\\
S3Net~\cite{yang2021s3net} & Depth & 0.512 & 0.331 & 0.418 & 0.374 \\
S3Net~\cite{yang2021s3net} & - & 0.499 & 0.336 & 0.414 & 0.377 \\
Latent-Intrinsic \cite{Latent-Intrinsic} ($\sigma = 0$) & - & 0.326 & 0.232 & 0.242& 0.541\\
Latent-Intrinsic \cite{Latent-Intrinsic} & - & 0.297 & 0.473& \textbf{0.222} & \textbf{0.571}\\
\midrule
\multicolumn{6}{l}{\textit{Trained on Diverse Indoor Images}} \\
RGB$\leftrightarrow$X \cite{zeng2024rgb} & G-Buffer & 0.587 & 0.070& 0.427 & 0.215\\
RGB$\leftrightarrow$X \cite{zeng2024rgb}(same scene) & G-Buffer & 0.523 & 0.134& 0.340 & 0.350\\
Ours (previous) & - & {0.343}& {0.359} & {0.267} & {0.448}\\
Ours (w/o bypass-decoder \cite{quadprior}) & - & 0.318 & 0.385 & 0.251 & 0.467\\
Ours (w/o seed selection) & - & 0.326 & 0.417 & 0.253 & 0.512\\
Ours & - & 0.310& 0.440 & 0.240 & 0.527\\
\bottomrule
\end{tabular}}
\vspace{-10pt}
\caption{\textbf{Quantitative Evaluation on MIIW.}
We evaluate our method on the MIIW dataset~\citep{MIIW}. The table is split into two panels.
The \textbf{top panel} shows specialist methods trained exclusively on MIIW. While these models perform well on this benchmark, they tend to overfit and generalize poorly to in-the-wild images.
The \textbf{bottom panel} shows generalist methods trained on more diverse datasets. Our method, trained on a diverse combination of data for better generalization, achieves strong performance competitive with the specialist models.
We note the known sensitivity of pixel-wise metrics~\cite{giroux2024towards}; for example, a one-pixel shift in a specular highlight can disproportionately lower scores without reflecting a change in perceptual quality. For this reason, we supplement this analysis with a human perceptual study in \cref{tab:user_study}. All metrics are reported without flow-based post-processing.
}
    \label{tab:MIIW}
    \vspace{-5pt}
\end{table}

\begin{table}[t!]
    \centering
    \resizebox{\linewidth}{!}{
    \begin{tabular}{l|c |c c c}
    \toprule
         \textbf{Method} &\multicolumn{1}{c}{\textbf{Surface Normal}}&\multicolumn{3}{|c}{\textbf{Perceptual Relighting Quality}}\\
         \midrule
          &Median-AE $\downarrow$ &{I-PQ} $\downarrow$& {L-PQ }$\downarrow$& {P-PQ}$\downarrow$   \\
        \midrule
        RGB$\leftrightarrow$X \cite{zeng2024rgb} & 3.14&2.21 &2.88 &2.70  \\
        IC-Light-v2 \cite{IC-Light}  &3.42 &3.06 &2.57 &2.74  \\
            Latent-Intrinsic \cite{Latent-Intrinsic}& 3.61 &2.24&2.52 &2.40  \\
        Ours& \textbf{2.74} &\textbf{1.71} &\textbf{1.30} &\textbf{1.40}  \\
        \bottomrule
    \end{tabular}}
    \vspace{-10pt}
        \caption{\textbf{Real-world Evaluation.} We evaluate surface normal consistency and conduct a user study inspired by \cite{kocsis2024lightit}, comparing our method against RGB$\leftrightarrow$X \cite{zeng2024rgb}, IC-Light-v2 \cite{IC-Light}, and Latent-Intrinsic \cite{Latent-Intrinsic}. Users rated images generated by each method under the same target lighting (image or text prompt) on image quality (I-PQ), lighting quality (L-PQ), and prompt alignment (P-PQ). Our approach outperforms all baselines across all metrics, demonstrating strong and robust open-world relighting capabilities.}
    \label{tab:user_study}
\end{table}

To evaluate the geometry consistency, we use RGB$\leftrightarrow$X~\cite{zeng2024rgb} to generate the surface normal for both the original and relight images and use the surface normal for the original image as the ground truth. Following the common evaluation protocol for the surface normal evaluation, we measure angular error (AE) for the pixels with ground truth and report the median value in Tab. \ref{tab:user_study}. Thanks to the carefully designed latent intrinsic condition, our method successfully preserves the geometry details with the median-AE lower than 3 degrees. While RGB$\leftrightarrow$X \cite{zeng2024rgb}, IC-Light-v2 \cite{IC-Light} and Latent-Intrinsic \cite{Latent-Intrinsic} all report error larger than 3 degrees. 

\subsection{Ablation Study}
\begin{table}[t]
\vspace{-3pt}
\begin{center}
\footnotesize
\begin{tabular}[t]{l|c|c}
\toprule
\textbf{Method} & \textbf{FID} $\downarrow$ & \textbf{CLIP-Score} $\uparrow$\\
\midrule
StyLitGAN & 47.986 & 28.948\\
Var. StyLitGAN & 37.068& 29.226\\
Var. StyLitGAN + CLIP filtering   & \textbf{35.810} & \textbf{29.288}\\
\bottomrule
\end{tabular}
\end{center}
\vspace{-4ex}
\caption{\textbf{Ablation for Variational-StylitGAN.} The variational (Var.) setting improves the quality of generated images, while CLIP filtering further enhances performance.}
\label{tab:FID}
\vspace{-15pt}
\end{table}

\begin{figure*}[t]
    \centering
    \includegraphics[width=\linewidth]{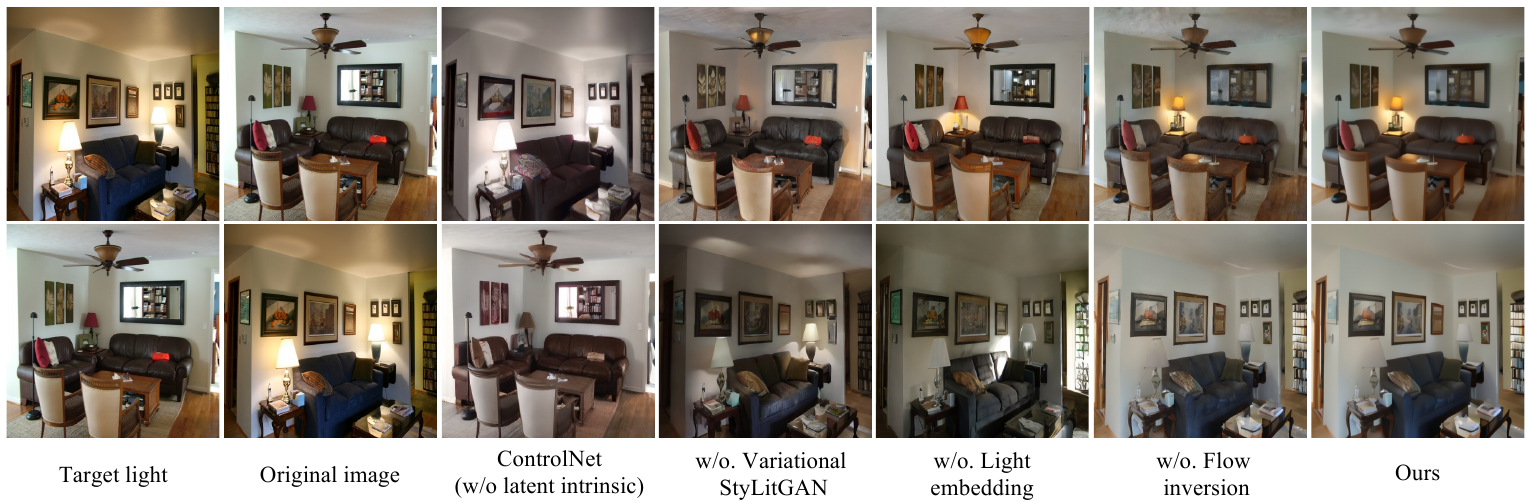}
    \vspace{-18pt}
    \caption{\textbf{Ablation Study}. Left: Target light. Second: Source image. Third: Vanilla ControlNet (without latent intrinsic) fails at relighting, altering average color while losing details. Fourth: Without variational StyleGAN data, \method fails to recognize light sources (e.g., switching lamps on/off). Fifth: Without the adaptor network and cross-attention fine-tuning, second-order effects like table gloss (top row) are lost. Sixth: Without flow inversion, relighting is reasonable but introduces artifacts. Last: Combining all components yields artifact-free, realistic relighting with second-order effects.}
    \vspace{-1.7ex}
    \label{fig:alblation}
\end{figure*}

We conduct a user study with 31 participants to access the perceptual image generation quality inspired by \cite{kocsis2024lightit}. The metrics are: 1) relit image quality (I-PQ), which evaluates the intrinsic preservation of the relit image; 2) lighting quality (L-PQ), which evaluates the realism of the lighting; and 3) alignment with the lighting prompt (P-PQ). The questions in the study included the original image, the target light image, and four randomly shuffled relit images (produced by the four aforementioned methods, respectively). For each metric, users are asked to rank the four relit images on a scale from one to four (where a lower score is better).  We can not compare with \cite{kocsis2024lightit} as it targets outdoor relighting, and their model is not publicly available. As reported in Tab. \ref{tab:user_study} we dominate the leaderboard by a notable margin, which again confirms the effectiveness of our method.

Quantitatively, Tab. \ref{tab:FID} presents the ablation results on the variational design and CLIP filtering for enhancing the stability and quality of StyLitGAN-generated images. The results indicate that the variational design improves FID by 22.7\% (47.9 → 37.0), while CLIP filtering provides an additional 3.4\% improvement (37.0 → 35.8).

\cref{fig:alblation} presents a visual ablation study. ControlNet \cite{zhang2023adding_controlnet} (2$^{nd}$ column) fails at meaningful relighting, producing an averaged color instead. Using the latent intrinsic condition (5$^{th}$–7$^{th}$ columns), the model learns lighting transfer, enabling effects like turning a lamp on/off. However, relying solely on this condition (5$^{th}$ column) fails to capture second-order effects (e.g., reflections), highlighting the need to fine-tune cross-attention layers. Flow-based inversion (7$^{th}$ column) reduces artifacts in \method. To assess Variational StyLitGAN's role in dataset generation, we removed its data. As shown in the 4$^{th}$ column, while general illumination can be learned, scene-specific lighting effects (e.g., lamps) are lost without paired relit images, emphasizing the importance of Variational StyLitGAN.  
\section{Discussion}

Our results show that photorealistic indoor scene relighting can be achieved through a purely image-based approach using latent representations. By designing latent intrinsic control within a diffusion-based framework, \method handles challenging lighting effects: including thin cast shadows, specular highlights, and indirect illumination; where previous methods struggle. Leveraging pretrained diffusion models and latent intrinsics, \method transfers lighting across diverse indoor scenes without relying on geometry or multi-view inputs.

Although trained only on paired samples from the same scene, \method generalizes to cross-scene relighting in the wild. It preserves scene structure and material while transferring lighting across scenes with different layouts and reflectance properties. This stems from architectural choices that combine latent intrinsics with conditional diffusion, enabling \method to learn generalizable lighting transfer beyond the training distribution.

While promising, several directions remain. Extending to dynamic scenes, improving 3D consistency, and achieving real-time inference are important next steps. Reducing artifacts without relying on post-hoc enhancements such as RF-Inversion also remains an open challenge. Nonetheless, the success of our latent-space approach points toward a broader shift in scene manipulation that does not depend on explicit physical modeling.

Recent work suggests that generative models struggle to model long-range lighting effects, often leading to shadow artifacts~\cite{sarkar2024shadows}. Our explicit lighting control indicates that some long-range phenomena can be learned. A key future direction is to systematically identify which classes of long-range behavior are captured and which remain out of reach.

\begin{figure}
    \centering
    \vspace{-5pt}
\includegraphics[width=0.95\linewidth]{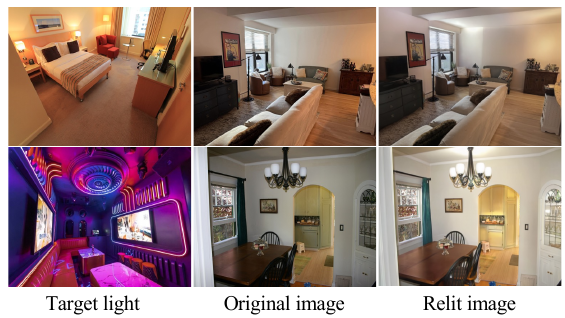}
\vspace{-8pt}
    \caption{\textbf{Failure case}. Our method fails to recognize the lamp when it is either too small or positioned with its back to the camera. It also fails to transfer dramatic lighting color (chromaticity), such as the lighting of a Karaoke room.}
    \label{fig:supp-bad}
    \vspace{-10pt}
\end{figure}
\noindent\textbf{Limitation.} We observe that \method struggles to detect small or occluded light sources, and fails to capture chromatic shifts in dramatic lighting (\cref{fig:supp-bad}). These issues may be alleviated by broader training data. Additionally, the model does not control light intensity. While relighting remains plausible, some outputs show discrepancies in brightness and color. Quantifying these deviations remains difficult due to the lack of ground truth. Our evaluation on Multi-Illumination confirms strong performance, though the dataset lacks broader scene complexity and dynamic lighting events like switching on and off visible luminaires.

\section*{Acknowledgment}
We are grateful to Tommy Mitchel, Takayuki Ishida and Shun Terasaki of the Sony PlayStation team for discovering a bug in our evaluation code. Their feedback was crucial for correcting the results in Table 1.

We thank Pingping Song, Melis \"{O}cal, and Alexander Timans for their insightful discussions. We are also grateful to David Forsyth for his suggestion to emphasize that light in scenes cannot simply appear, along with other valuable comments on our manuscript. Additionally, we thank Xiao Zhang for providing the code and model for Latent Intrinsic. Finally, we thank all our participants for their time in finishing our user study. This project is financially supported by Bosch (Bosch Center for Artificial Intelligence), the University of Amsterdam and the allowance of Top consortia for Knowledge and Innovation (TKIs) from the Netherlands Ministry of Economic Affairs and Climate Policy.

{
    \small
    \bibliographystyle{ieeenat_fullname}
    \bibliography{main,bibs/bib_mani,bibs/bigdaf,bibs/dafjulia,bibs/dafsupp,bibs/jasonrefs,bibs/lana,bibs/references,bibs/refs,bibs/in_GS,bibs/retinexdiff}
}
\clearpage
\setcounter{page}{1}
\renewcommand{\thefigure}{S.\arabic{figure}}
\setcounter{figure}{0} %
\maketitlesupplementary
\appendix
\noindent\textbf{Outline.} We begin by detailing the training setup and implementation, followed by additional relit results and an illustration of the nearest-neighbor search.

\section{Training details}
We use Stable Diffusion 2.1 \cite{rombach2022high_latentdiffusion_ldm} as our base model to balance performance and training costs. To better preserve the details of the input images, we jointly estimate the de-noised image and noise map at each denoising step (known as the \( v \)-prediction). Our method also applies to other objective functions, such as \( \epsilon \) (only predicts the noise map). All training and testing are conducted on an 8-GPU NVIDIA A6000 Ada 48GB node. For the SD2.1 base model, we train on images with a resolution of 512 $\times$ 512. An AdamW \cite{loshchilov2017decoupled} optimizer with a learning rate of \(4 \times 10^{-5}\) and a decay rate of 0.9 is used. Training requires approximately 120 hours on a single GPU. At inference time, \method outputs a relighted image (resolution: 512 $\times$ 512) in 5 seconds with 50 DDIM steps.

\begin{figure*}[t!]
    \centering
    \includegraphics[width=\linewidth]{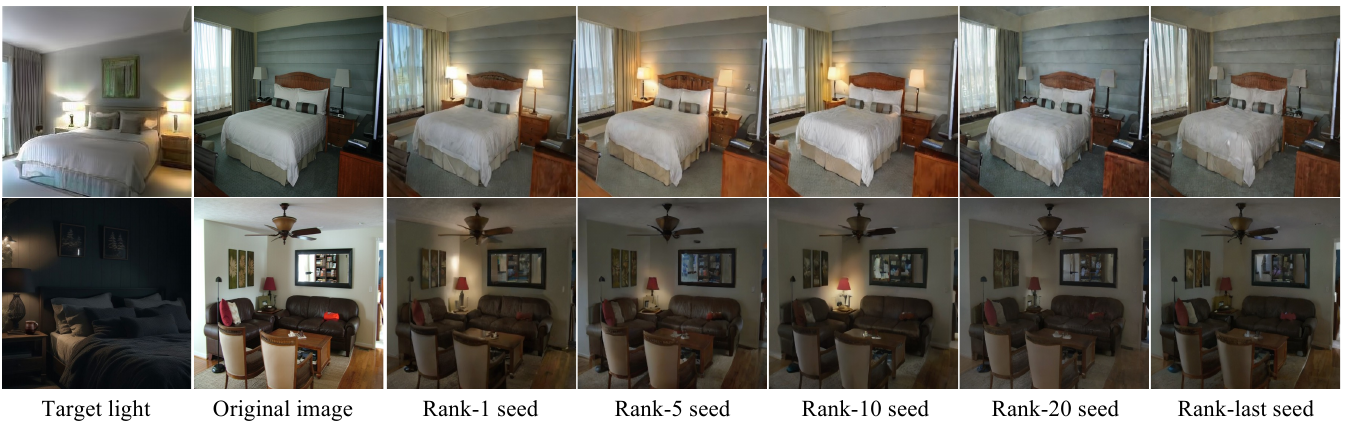}
    \vspace{-18pt}
    \caption{\textbf {Nearest Neighbor Search.} Diffusion models are sensitive to seed choice~\cite{xu2024good}. We observed that the choice of random seeds significantly impacts relighting quality. Here, we present sampled relights generated from 30 random seeds, sorted by their match to the target lighting image. Sorting is based on nearest-neighbor matching of the latent extrinsic (a low-dimensional lighting vector) to the target.
    }
    \label{fig:seed}
    \vspace{-1.5ex}
\end{figure*}

\section{More results}
\paragraph{Nearest neighbor search}
In complex real-world scene relighting, particularly for fine-grained light control, we observe that light transfer is highly sensitive to the choice of seed. To address this issue, we propose a nearest neighbor search over multiple seed candidates to identify the relit result that best approximates the target lighting.

\cref{fig:seed} illustrates the nearest neighbor search over multiple seed candidates to identify the relit result that best approximates the target lighting.

\begin{figure*}[t!]
    \centering
    \includegraphics[width=0.9\linewidth]{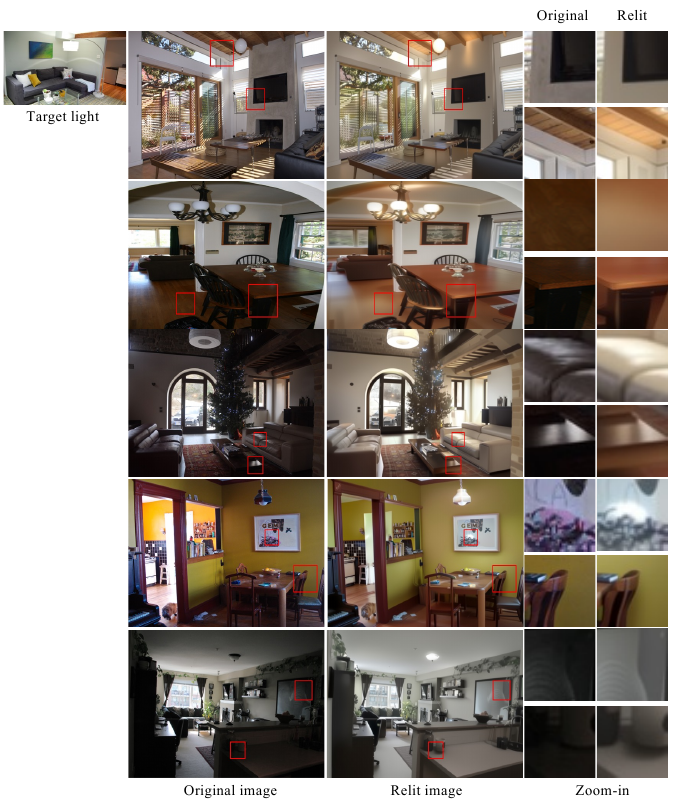}
    \caption{\textbf{Additional Relit Images (switching on ceiling lamps)}. The target lighting is shown in the top-left image, where a ceiling lamp is switched on. Ceiling lamps are very rare in our training data; however, we find that \method is still able to understand them and synthesize plausible relit images, as shown in the third column. In the first row, notice the suppression of gloss near the window at the top (see crop) and the added gloss due to inter-reflection on the TV screen. In the second row, observe how three ceiling lamps significantly brighten the room, with strong gloss visible on both the wooden floor and the dining table. In the third row, notice the sheen on the sofa and the edge of the coffee table, which become clearly visible after relighting. In the fourth row, see how the reflection of the lamp appears on the painting on the side wall. Also, note the shadow cast by the chair on the side wall below the painting. Finally, in the last row, observe the reflection on the mirror-like surface and note the cast shadow of the coffee cup visible in the bottom crop.}
    \label{fig:supp-celing}
\end{figure*}

\begin{figure*}[t!]
    \centering
    \includegraphics[width=0.9\linewidth]{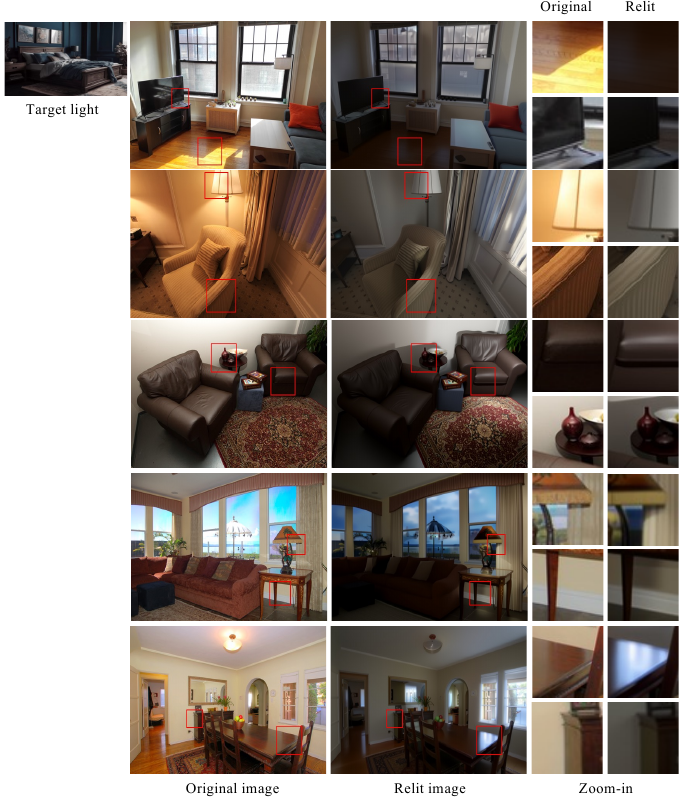}
    \caption{\textbf{Additional Relit Images}. The target light is shown in the top-left image, where all lamps are switched off, and the only illumination comes from diffused natural light entering through a window on the right. The second column displays the source images to be relit to match the target light, while the third column presents the relit images. The final column highlights cropped regions before and after relighting, emphasizing the second-order lighting effects captured by \method. In the top row (first relit image), note the table's reflection in the TV and the strong gloss on the table from the directional window light. In the fourth row, observe how the sky changes to reflect the ambiance of the target light. In the last row, notice specular highlights on the table because of the direction light from the window. Also, notice the shadow cast by the cabinet in the bottom crop.}
    \label{fig:supp-more3}
\end{figure*}

\begin{figure*}[t!]
    \centering
    \includegraphics[width=0.9\linewidth]{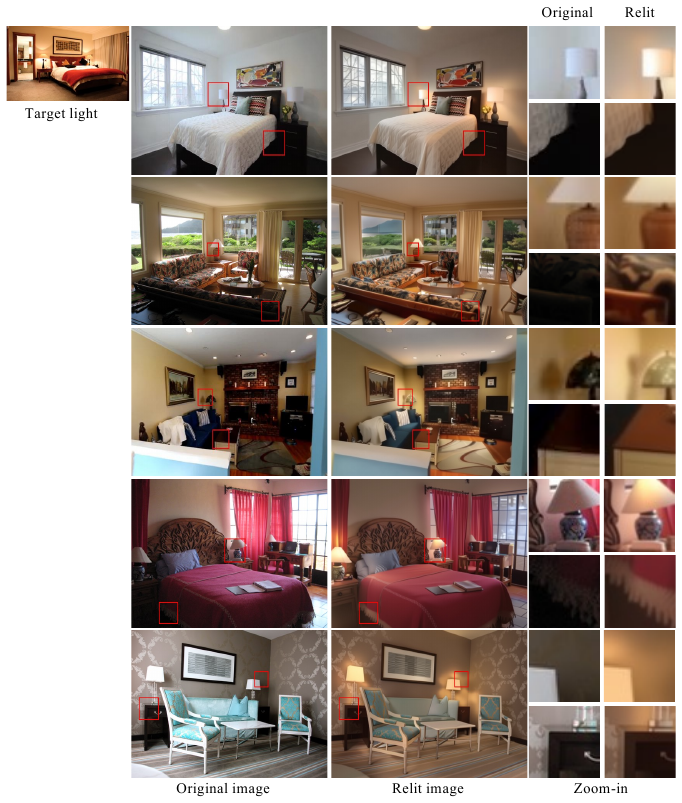}
    \caption{\textbf{Additional Relit Images}. The target lighting is shown in the top-left image, where all lamps are switched on. The second column displays the source images to be relit to match the target lighting, where all lamps are switched off, and the third column presents the relit images. The final column highlights cropped regions before and after relighting. In the top row (first relit image), note the overall change in the room's color and the colored gloss added to the side of the bedsheet. In the second row, notice that the strong gloss on the carpet is removed. In the third row, switching on the side lamps removes the lamp shadow; also, observe the effect of the lamp on the ceiling and the gloss added to the edge of the table, as shown in the crop. In the fourth row, notice that the left side of the bed is now well-lit due to the lamp. Finally, in the last row, observe the gloss added to the wallpaper because of switching on the lamp}
    \label{fig:supp-more_on}
\end{figure*}

\paragraph{Additional relit images}
We present additional relit results in the following pages to demonstrate the robustness of our method under varying lighting conditions and across diverse scenes. These include a scenario showcasing the effect of turning on ceiling lights (\cref{fig:supp-celing}), an example illustrating reduced ambient lighting in the room (\cref{fig:supp-more3}), and results depicting the effect of turning on lamps (\cref{fig:supp-more_on}). We provide a detailed analysis of these phenomena in the figure captions and highlight the lighting effects using red bounding boxes within the figures.

\end{document}